\documentclass{article}

\usepackage[final]{neurips_2023}

\usepackage[utf8]{inputenc} 
\usepackage[T1]{fontenc}    
\usepackage{nameref, hyperref}       
\usepackage{url}            
\usepackage{booktabs}       
\usepackage{amsfonts}       
\usepackage{nicefrac}       
\usepackage{microtype}      
\usepackage{xcolor}         
\usepackage{float}
\usepackage{graphics}
\usepackage{graphicx}
\usepackage{etoolbox}
\usepackage{amsmath}
\usepackage{makecell}

\usepackage{xcolor}

\title{Analyzing the temporal dynamics of linguistic \\ features contained in misinformation }

\author{%
  Erik J.~Schlicht \\
  Misinformation-Monitor\\
  \texttt{misinfo-monitor.org} \\
  \texttt{erik@misinfo-monitor.org} \\
}

\begin{document}

\maketitle

\begin{abstract}
Consumption of misinformation  can lead to negative consequences that impact the individual and society.  To help mitigate the influence of misinformation on human beliefs, algorithmic labels providing context about content accuracy and source reliability have been developed. Since the linguistic features used by algorithms to estimate information accuracy can change across time, it’s important to understand their temporal dynamics.  As a result, this study uses natural language processing to analyze PolitiFact statements spanning between 2010 and 2024 to quantify how the sources and linguistic features of misinformation change between five-year time periods.  The results show that statement sentiment has decreased significantly over time, reflecting a generally more negative tone in PolitiFact statements.  Moreover, statements associated with misinformation realize significantly lower sentiment than accurate information. However, the difference in sentiment between misinformation and accurate information has recently diminished due to decreased sentiment associated with accurate statements.  Additional analysis shows that recent time periods (2020-2024) are dominated by sources from online social networks (e.g., X, Facebook) and other digital forums (e.g., Blogs, Viral Image) that contain high levels of misinformation containing negative sentiment.  In contrast, most statements during early time periods (2010-2014) are attributed to individual sources (i.e., politicians) that are relatively balanced in accuracy ratings and contain statements with neutral or positive sentiment.  Named-entity recognition was used to identify that presidential incumbents and candidates are relatively more prevalent in statements containing misinformation, while US states tend to be present in accurate information.    Finally, entity labels associated with people and organizations are more common in misinformation, while accurate statements are more likely to contain numeric entity labels, such as percentages and dates.          
 \end{abstract}

\section{Introduction}
Misinformation is a statement that contains false or misleading information and has resulted in negative consequences to individuals and society.  Individuals who consumed misinformation surrounding the COVID pandemic were more likely to demonstrate vaccination hesitancy~\cite{1,2} and experience mental health consequences~\cite{3}.  The societal impact of misinformation includes distrust of the media~\cite{4}, erosion of civil discourse, and political paralysis~\cite{5}.  Despite its serious impact on individuals and society, misinformation is known to be shared more than valid information~\cite{6}, and the reasons are diverse and include cognitive factors~\cite{7,8}, socio-affective factors~\cite{8}, incentives~\cite{9}  and changes in the information system~\cite{5,10}.

Given its potential for harm and propensity to spread, there have been several methods proposed to mitigate the impact of misinformation~\cite{11}.   Promising approaches include supporting local journalism~\cite{12,13}, media literacy and education~\cite{14,15}, and modifying recommendation algorithms~\cite{16,17}.   However, the first two solutions suffer from limited scalability, and the latter has not been thoroughly investigated~\cite{11}.  Making matters worse, online social networks have recently abandoned third-party fact-checking~\cite{18,19}, thereby removing an approach that has shown reasonable effectiveness and scalability~\cite{11,20,21}.  

A highly scalable alternative involves labeling social media content to provide context about the reliability of the source and content accuracy~\cite{11,22,23}.   Although current interventions have demonstrated mixed effectiveness~\cite{24,25}, identifying robust labels that reduce the impact of misinformation would provide a scalable solution.   Since external fact-checking is no longer being used by major platforms~\cite{18,19}, it will require algorithmic solutions to produce labels for information consumers.  Moreover, automated labels must realize a high level of accuracy in order to elicit trust from the user. 

Algorithms that estimate the reliability of sources have taken many forms and are capable of estimating if the source is a bot~\cite{26,27}, spammer~\cite{28,29}, or a fake profile~\cite{30,31}, while another uses reinforcement learning to estimate the overall reliability of the source~\cite{32}.  Natural language processing (NLP) is often used to quantify linguistic features of text-based claims that are used to estimate content accuracy~\cite{33, 34}.  Since misinformation relies on emotional content, such as appealing to morality and statements with negative sentiment~\cite{35}, sentiment analysis is sometimes leveraged by NLP-based approaches~\cite{36, 37}.  Named-entity recognition (NER) is also used to identify entities contained in text that are predictive of misinformation~\cite{38, 39}.   

Since the linguistic features used to estimate source reliability and content accuracy can change across time, it’s important to understand their temporal dynamics.  As a result, this study uses NLP to analyze PolitiFact statements spanning between 2010 and 2024 to quantify how features of misinformation change between five-year time periods. Other studies have investigated linguistic properties of misinformation, but used a narrow time range in their investigation~\cite{40,40b} or AI-generated content~\cite{41}. Moreover, this paper extends previous work investigating the temporal dynamics of misinformation by directly contrasting how linguistic features change between time periods~\cite{42}.  These insights can be used by developers to inspire algorithmic solutions to mitigate the impact of misinformation.  The next section will detail the data and methods used for this investigation.

\section{Methods}
\label{methods}
Misinformation was identified by leveraging fact-checked claims from \href{https://www.politifact.com} {PolitiFact}, which is owned by the nonprofit  \href{https://www.poynter.org/about} {Poyneter Institute for Media} with the objective of improving the relevance, ethical practice and value of journalism.  As part of that objective, PolitiFact verifies the accuracy of claims made throughout various online and media sources.  Fact-checked information provides a source of ground-truth for information accuracy with relatively high confidence.  For the purpose of this investigation, the term misinformation is used generally to include both misinformation and disinformation since the intent behind false information on PolitiFact is not observable.  

Approximately twenty-four thousand (23,786) PolitiFact samples were collected across years ranging between 2010 and 2024.  Each sample contained metadata about the year of the post, the source of the claim, a statement that summarizes the fact-checked claim, and the fact-check rating that was further classified into one of three rating type categories: 
\begin{itemize}
  \item \textbf{Accurate:}  this rating type was given to fact-checked ratings with TRUTH-O-METER scores of  TRUE or MOSTLY-TRUE.
  \item  \textbf{Misinfo:}  this rating type was given to fact-checked ratings with TRUTH-O-METER scores of  PANTS-ON-FIRE or FALSE.
  \item  \textbf{Mixed:}  this rating type was given to fact-checked ratings with all other TRUTH-O-METER ratings not contained in the classes above.
  \end{itemize}
  
\begin{figure}[h]
  \centering
  \includegraphics[width=14cm]{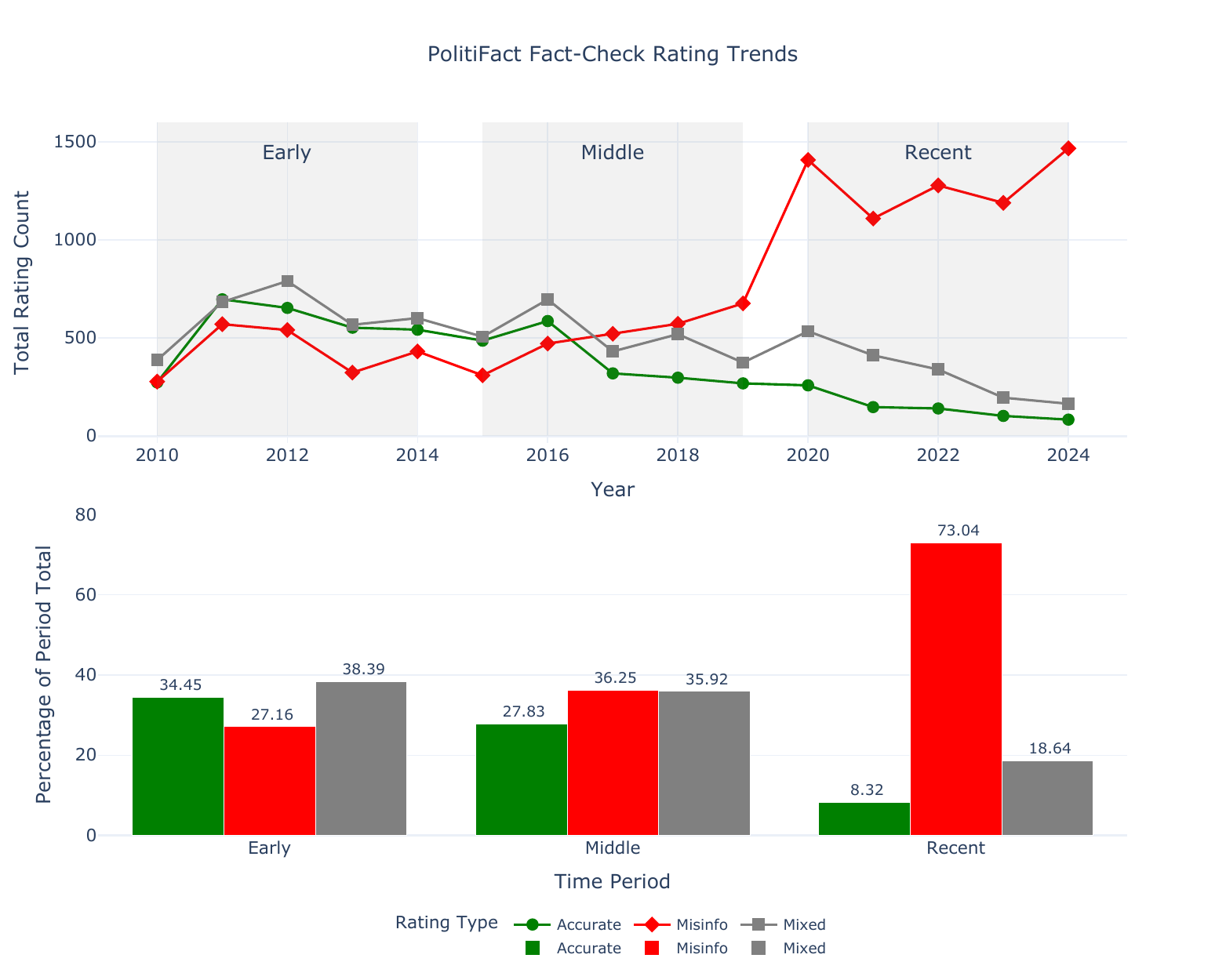}
  \caption{PolifiFact rating type trends (top) and rating type distributions within time periods (bottom).}
   \label{fig1}
\end{figure}

Rating type categories were selected to facilitate analysis and plotting and to ensure that accurate and misinformation categories are associated with statements that reflect high (Accurate) or low (Misinfo) levels of accuracy.  Moreover, years were evenly binned into 5 year time periods that correspond Early (2010-2014), Middle (2015-2019) and Recent (2020-2024) time periods (Figure~\ref{fig1}).  
 
As Figure~\ref{fig1} shows, rating type distributions for PolitiFact statements are approximately uniform  across early and middle time periods, followed by a recent spike in misinformation beginning in 2020.  It's interesting to note that this was near the onset of the COVID-19 Infodemic~\cite{43}.   When considering the full range of years (2010-2024), approximately 47\% of statements are associated with misinformation, whereas the remainder contained either mixed ($\sim$30\%) or accurate ($\sim$23\%) information. PolitiFact data was further analyzed to produce the results in the next section.  All of the data and analysis scripts used to produce the figures and tables in this paper can be found in supplementary files ( \nameref{S1_File}).

\section{Results}
\label{results}
This section details results from the analysis performed using the data and methods described in the last section.  Since the linguistic features used to produce automated labels can change across time, it’s important to understand their temporal dynamics.  As a result, this section explores features that are commonly used in automated approaches~\cite{36, 37, 38, 39}.  First, sentiment analysis is performed on the PolitiFact statements to investigate how the emotional content changes across time periods and rating types.  Next, analysis is performed to provide insights into common sources of misinformation and the consistency of those sources across time periods.  Finally, the last two sections use NER to discover entities and entity labels that are common in misinformation, and also explores their consistency across time periods.  

\subsection{Statement Sentiment Trends}
Investigating sentiment is important since previous research found that misinformation relies on emotional content, such as appealing to morality and statements with negative sentiment~\cite{35}, and automated approaches to detecting misinformation have also leveraged sentiment~\cite{38, 39}. Therefore, this section explores sentiment associated with PolitiFact statements by leveraging the compound score produced by VADER~\cite{44}.  This score ranges between -1 and +1, with scores less than 0 corresponding to a statement with negative sentiment, scores greater than 0 corresponding to statements with positive sentiment, and scores around 0 corresponding to neutral statements. 

Since the original claims/posts are not always available and span information modalities (text, image), sentiment analysis was performed on PolitiFact statements. These statements provide a summary of the claim being fact-checked, allowing the sentiment associated with the original content to be estimated. Although this is a sanitized version of the original, it provides a consistent and feasible method to measure content sentiment.  That said, this is a limitation associated with this investigation, as it may impact the generalizability of these findings to the original content. 

\begin{figure}[t]
  \centering
  \includegraphics[width=14cm]{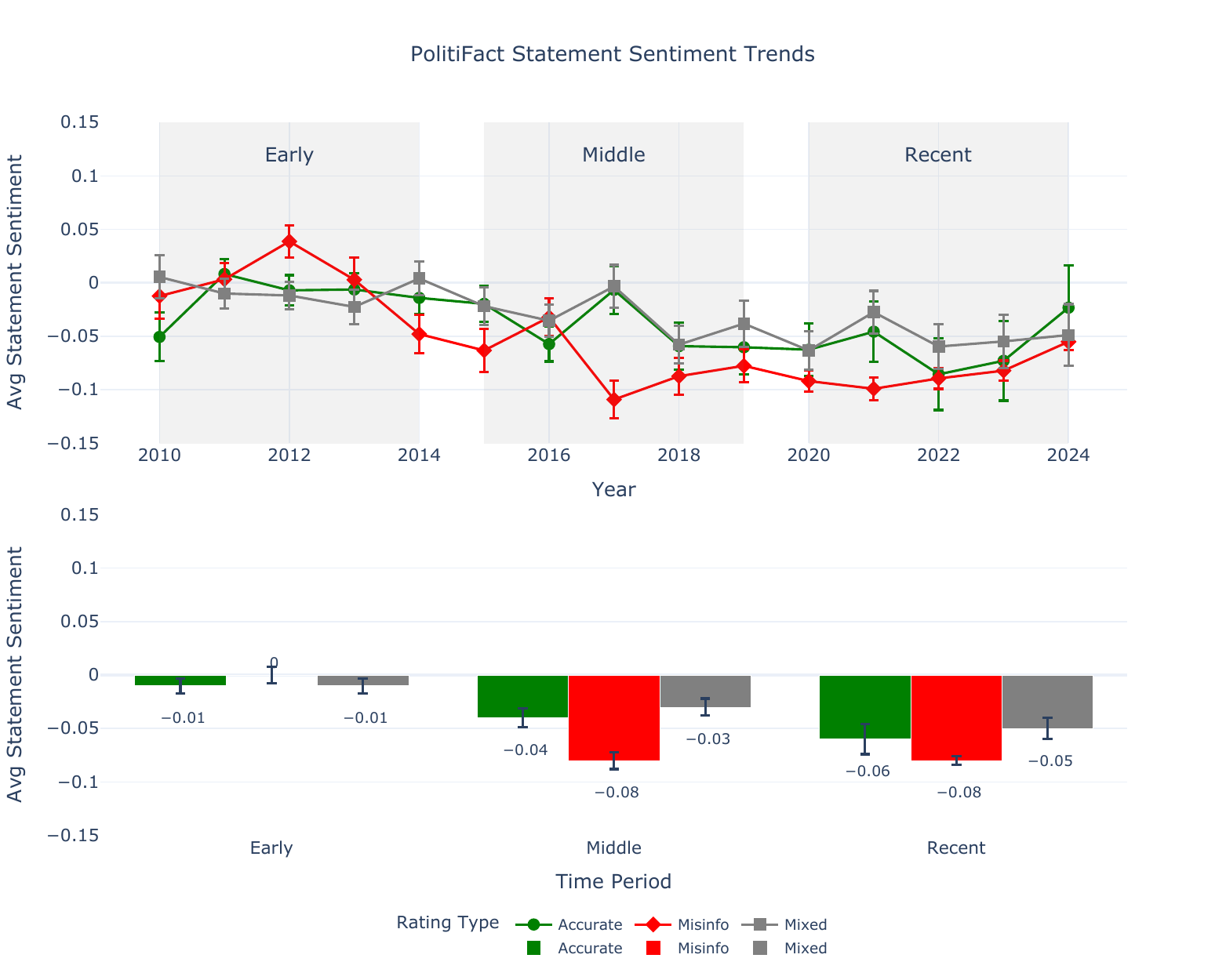}
  \caption{PolifiFact statement sentiment trends (top) and average sentiment within time periods (bottom).}
   \label{fig2}
\end{figure}

The top panel of Figure~\ref{fig2} shows mean sentiment for each year ($\pm$ 1 SEM) and rating type (color).   The gray rectangles reflect the five-year time periods used in this study across early (2010-2014), middle (2015-2019) and recent (2020-2024) periods.   General trends show that the sentiment associated with the early time period appears to be relatively neutral (near 0) with little difference in sentiment between rating types, whereas sentiment associated with middle and recent time periods are slightly negative, with misinformation (red) realizing lower average sentiment than accurate (green) and mixed (gray) ratings for many of the years in those periods. 

The bottom panel of Figure~\ref{fig2} shows mean sentiment within a time period ($\pm$ 1 SEM) and rating type (color).  In order to directly compare conditions, a 3 (Period) x 3 (Rating Type) Aligned Rank Transform (ART) ANOVA was performed~\cite{45}.  The parametric version was not used since its assumptions are violated.  The results show that there are significant main effects of both of time period (F(2, 23777) = 60.78, p<.001) and rating type (F(2, 23777) = 43.96, p<.001), in addition to a significant interaction (F(4, 23777) = 2.80, p<.05).

A Mann-Whitney U-Test with Bonferroni correction was performed to compare pairwise differences.  The results show a significant decrease in sentiment across all time periods, with early periods (M = -0.006, SEM = 0.004) displaying significantly greater sentiment (U = 29,552,972, p<.001) than middle periods (M = -0.050, SEM = 0.005), which have significantly greater sentiment (U = 32,208,867, p<.001)  than recent periods (M = -0.075, SEM = 0.004).  Note that recent time periods also realize significantly lower sentiment than early time periods (U = 38,475,297, p<.001).  This result suggests a generally more negative tone associated with PolitiFact statements over time periods. 

Additional analysis demonstrates that misinformation (M = -0.065, SEM = .003) has significantly lower sentiment (U = 31,977,804, p<.001) than accurate statements (M = -0.065, SEM = .003), and significantly lower sentiment (U = 38,005,407, p<.001) than mixed-accuracy statements  (M = -0.026, SEM = .005).  However, the sentiment for accurate and mixed-accuracy statements are statistically identical (U = 19,564,607, p$\gg$.05).  This further supports research suggesting that misinformation relies on emotional content~\cite{35} .

Pairwise analysis of the interactions shows a significant decrease in statement sentiment associated with accurate information between early and recent periods (U = 1,074,489, p<.05) but not other periods.  Moreover, sentiment associated with misinformation in the early period was significantly greater than both the middle (U = 3,042,479, p<.001) and recent (U = 7,804,999, p<.001) time periods.  Taken together, these two findings suggest that statement sentiment may now be a less reliable signal of misinformation than it was during middle time periods.   Finally, sentiment associated with mixed-accuracy statements significantly decreased from early to recent periods (U = 2,640,529, p<.05), but otherwise realized similar sentiment across time periods (Figure~\ref{fig2}).   The next section will investigate the sources associated with PolitiFact statements. 

\subsection{Statement Source Analysis}
A highly scalable option for mitigating the impact of misinformation involves labeling social media content to provide context regarding the reliability of the sources~\cite{11,22,23}.  As a result, it is important to understand how the sources associated with misinformation change across time.  Therefore, this section analyzes the similarity of PolitiFact statement sources across time periods. 

\begin{figure}[h]
  \centering
  \includegraphics[width=14cm]{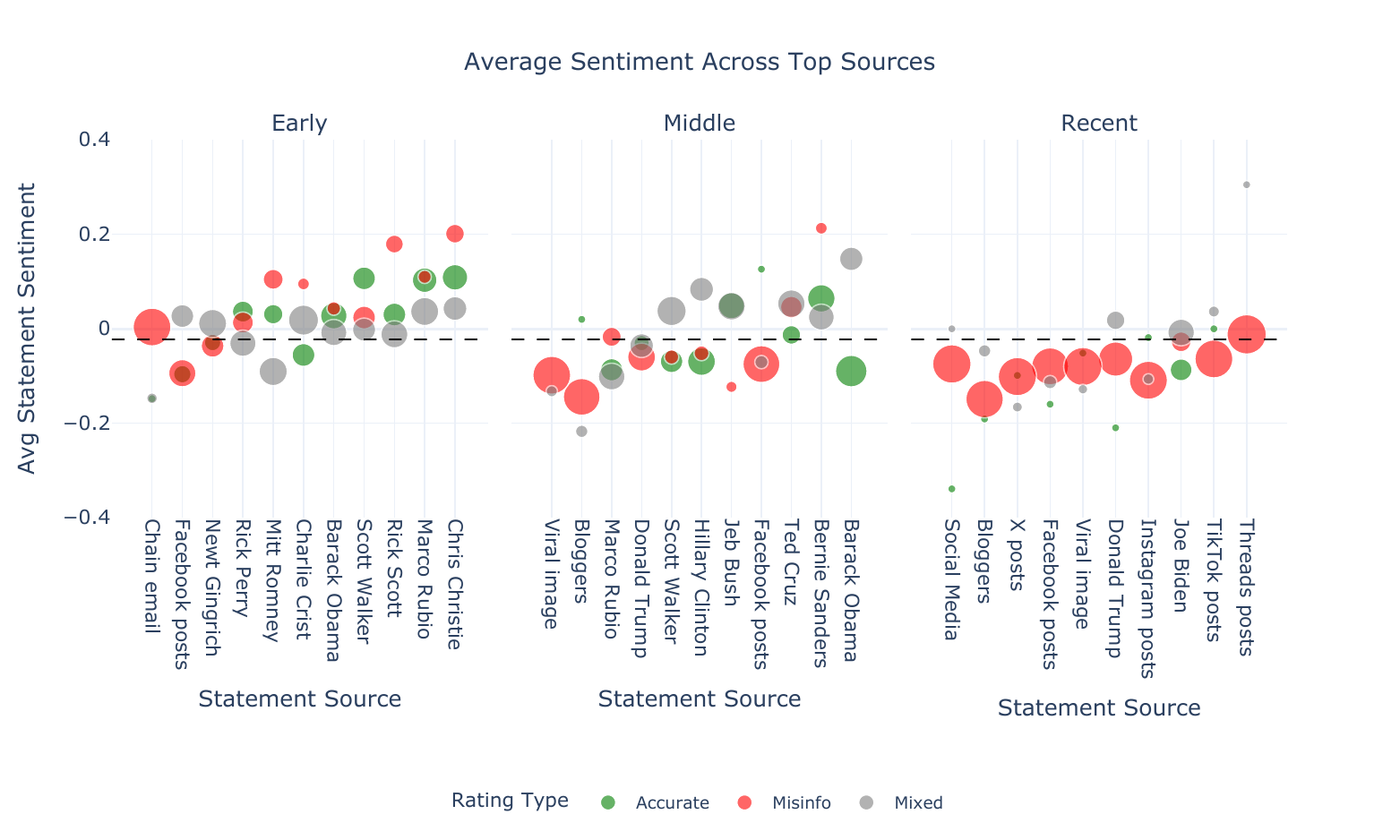}
  \caption{Top Sources of PolitiFact statements across time periods. }
   \label{fig3}
\end{figure}

Figure ~\ref{fig3} shows mean sentiment (y-axis) calculated within each PolitiFact statement source (x-axis) and rating type(color)  to investigate the accuracy and sentiment associated with different sources. The size of the marker is proportional to the number of samples within each point and the x-axis is sorted by the total number of samples for each source. The black dashed line represents average sentiment ($\sim$ -.05) across all PolitiFact statements.  Note that only the top 10 sources within each time period are plotted to avoid clutter. 

Online social networks (X, Facebook, Instagram, TikTok) and digital sources (Viral image, Bloggers) tend to be associated with high levels of misinformation (large red bubbles) and negative sentiment (below dashed black line), whereas sources attributed to individual politicians tend to be more balanced across rating types and contain neutral or positive sentiment.  Since online social networks and digital sources were not as prevalent in top sources for the early time period ($20\% $), source sentiment within this time period is greater (above dashed line), than recent time periods where these sources are more common ($80\% $).  

To quantify similarity in PolitiFact statement sources associated with varying levels of accuracy across time periods, sources ($s$) within each time period ($p$) were ranked separately by the total number of times the source was present in misinformation  ($R_{p,s}^{m}$) and accurate information ($R_{p,s}^{a}$).  Ranked Biased Overlap (RBO) similarity was computed on the ranked sources to evaluate source similarity across time periods for both misinformation and accurate information~\cite{46}.  An RBO score ranges from 0 (completely different sources) to 1 (identical sources).  Note that the RBO method allows sources to change across time periods and a weighting factor was used that gives $80\% $ the similarity score to the top 10 sources.  The results from this analysis are shown in Table ~\ref{table1}.

\begin{table}[h]
\small
  \caption{Source Similarity Across Rating Type and Time Periods.}
  \label{class-table}
  \centering
  \begin{tabular}{lll}
    \toprule
    \multicolumn{3}{c}{Source Similarity Scores }      \\
    \cmidrule(r){1-3}
    Rating Type  & Period Contrast & Similarity   \\
    \midrule
    Accurate  & Early-Middle & 0.19   \\
     ($R_{p,s}^{a}$) & Middle-Recent & 0.40  \\
     & Early-Recent & 0.06  \\
    \midrule
      Misinfo & Early-Middle & 0.19   \\
      ($R_{p,s}^{m}$) & Middle-Recent & 0.39  \\
     & Early-Recent & 0.07  \\
    \bottomrule
  \end{tabular}
  \label{table1}
\end{table}

Source similarity between middle and recent time periods is moderate (Accurate = .40, Misinfo = .39), whereas it is low between early and middle time periods  (Accurate = .19, Misinfo = .19), and almost completely different when comparing early and recent time periods  (Accurate = .06, Misinfo = .07).   There is also a high level of agreement in source similarity between time periods for misinformation and accurate information.  This suggests that top PolitFact statement sources for both rating types have vastly changed since early time periods and have become increasingly similar in recent time periods (Table ~\ref{table1}).  The next section will investigate common entities associated with PolitiFact statements.

\subsection{Statement Entity Analysis}
Named-entity recognition is  used to identify entities contained in text that are predictive of misinformation~\cite{38, 39}.  As a result, it is important to understand how the entities associated with misinformation change across time.  To that end, entities ($e$) within each time period  ($p$) were ranked separately by the total number of times the entity was present in misinformation ($R_{p,e}^{m}$) and accurate information ($R_{p,e}^{a}$).  Entities and entity labels were identified in PolitiFact statements using \href{https://spacy.io/} {spaCy}.  

\begin{figure}[h]
  \centering
  \includegraphics[width=14cm]{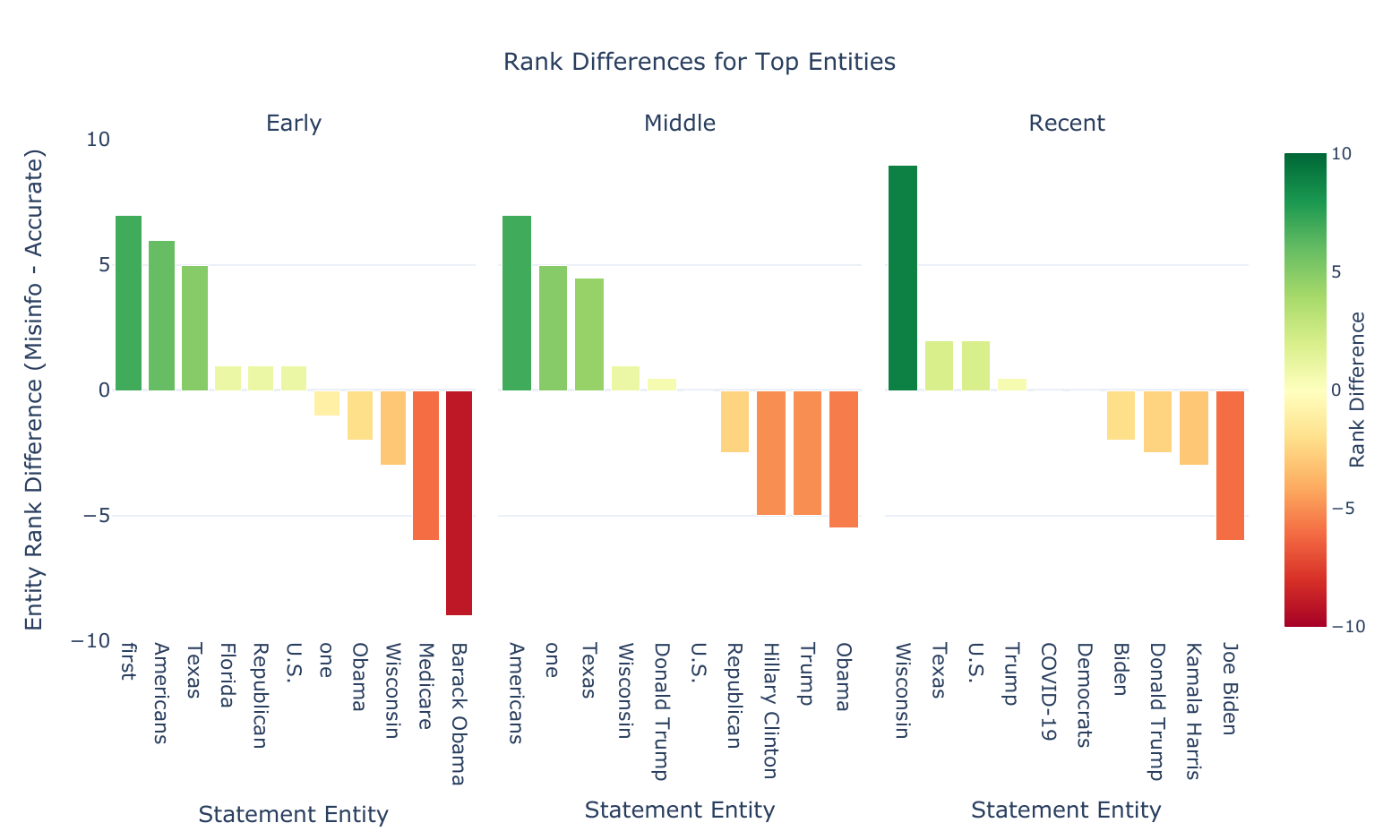}
  \caption{Top entities contained in PolitiFact statements across time periods.}
   \label{fig4}
\end{figure}

In Figure~\ref{fig4}, top entities contained in PolitiFact statements are presented along the x-axis and differences in the rating type ranks ($R_{p,e}^{m}$ - $R_{p,e}^{a}$) are plotted on the y-axis.   As a result, entities on the left-side of the chart are present in accurate statements more than misinformation (green), whereas entities on the right-side of the graph are found more in misinformation (red).  Note that only the top 10 entities contained in statements are plotted to avoid clutter.   With some exceptions, it appears that presidential incumbents and candidates are relatively more prevalent in statements containing misinformation, while US states tend to be present in accurate information.

RBO similarity was again computed on the ranked entities to evaluate entity similarity across time periods for both misinformation and accurate information (Table ~\ref{table2}).  Entity similarity is moderate between middle and recent time periods is moderate (Accurate = .46, Misinfo = .44), and low when comparing early and recent time periods  (Accurate = .24, Misinfo = .23).  Interestingly, entities associated with accurate information realize moderate similarity between early and middle time periods (.44), but entities associated with misinformation have lower similarity (.23).   
 
\begin{table}[h]
\small
  \caption{Entity Similarity Across Rating Type and Time Periods.}
  \label{class-table}
  \centering
  \begin{tabular}{lll}
    \toprule
    \multicolumn{3}{c}{Entity Similarity Scores }      \\
    \cmidrule(r){1-3}
    Rating Type  & Period Contrast & Similarity   \\
    \midrule
    Accurate  & Early-Middle & 0.44   \\
     ($R_{p,e}^{a}$) & Middle-Recent & 0.46  \\
     & Early-Recent & 0.24  \\
    \midrule
      Misinfo & Early-Middle & 0.23   \\
      ($R_{p,e}^{m}$) & Middle-Recent & 0.44  \\
     & Early-Recent & 0.23  \\
    \bottomrule
  \end{tabular}
  \label{table2}
\end{table}

This suggests that PolitFact statement entities associated with misinformation during early time periods have less similarity to those in other time periods.  Moreover, entities associated with accurate information have increased in similarity after the early time period.  The next section will investigate common entity labels associated with PolitiFact statements.

\subsection{Entity Label Analysis}
Since the total number of unique entities contained in the PolitiFact statements for this study exceeds fourteen-thousand, this section explores the eighteen categorical labels associated with statement entities.  This allows investigation into how entity labels associated with PolitiFact statements changes across time periods.  To that end, entity labels  ($l$) within each time period  ($p$) were ranked separately by the total number of times the entity was present in misinformation ($R_{p,l}^{m}$) and accurate information ($R_{p,l}^{a}$).  

\begin{figure}[h]
  \centering
  \includegraphics[width=14cm]{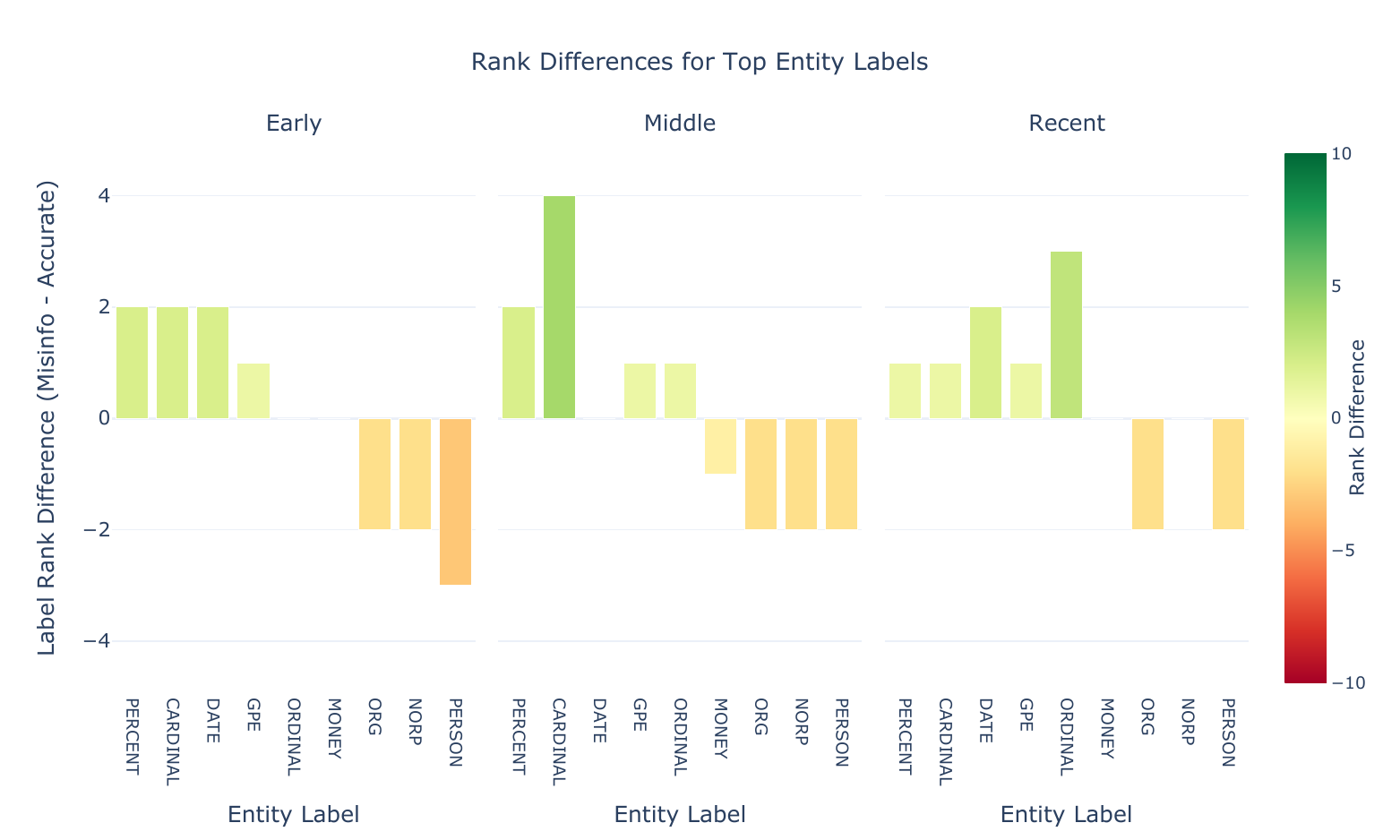}
  \caption{Top entity labels contained in PolitiFact statements across time periods.}
   \label{fig5}
\end{figure}

In Figure~\ref{fig5}, top entity labels contained in PolitiFact statements are presented along the x-axis and differences in the rating type ranks ($R_{p,l}^{m}$ - $R_{p,l}^{a}$) are plotted on the y-axis.   As a result, labels on the left-side of the chart are present in accurate statements more than misinformation (green), whereas labels on the right-side of the graph are found more in misinformation (red).   It appears that entity labels associated with people (PERSON) and organizations (ORG) are more common in misinformation, while accurate statements are more likely to contain numeric entity labels, such as percentages (PERCENT) and dates (DATE). Note that the differences at the entity label level (Figure~\ref{fig5}) tend to be smaller than at the raw entity level (Figure~\ref{fig4}).  Therefore, using entity labels will likely result in decreased ability to detect misinformation, but leveraging entities may require added processing due to the increased number of categorical features.  

Ranked Biased Overlap similarity was computed on the ranked entity labels to evaluate label similarity across time periods for both misinformation and accurate information (Table ~\ref{table3}).  Since there are fewer labels than entities, the RBO weighting factor used gives $80\% $ the similarity score to the top 5 sources.   

\begin{table}[h]
\small
  \caption{Entity Label Similarity Across Rating Type and Time Periods.}
  \label{class-table}
  \centering
  \begin{tabular}{lll}
    \toprule
    \multicolumn{3}{c}{Entity Label Similarity Scores }      \\
    \cmidrule(r){1-3}
    Rating Type  & Period Contrast & Similarity   \\
    \midrule
    Accurate  & Early-Middle & 0.79   \\
     ($R_{p,l}^{a}$) & Middle-Recent & 0.95  \\
     & Early-Recent & 0.81  \\
    \midrule
      Misinfo & Early-Middle & 0.79   \\
      ($R_{p,l}^{m}$) & Middle-Recent & 0.95  \\
     & Early-Recent & 0.81  \\
    \bottomrule
  \end{tabular}
  \label{table3}
\end{table}

Entity labels for PolitiFact statements show reasonably high similarity between all time periods.   There is also a perfect agreement in label similarity between time periods for misinformation and accurate information.   This suggests that top PolitFact statement entity labels (Table ~\ref{table3}) are generally more consistent than top entities (Table ~\ref{table2}) across time periods, but as discussed above, may result in decreased ability to detect misinformation. 

\section{Conclusion}

This study used NLP to quantify how features of misinformation change between five-year time periods, focusing on features that are commonly used in automated misinformation labeling approaches~\cite{36, 37, 38, 39}.   First, sentiment analysis was performed on the PolitiFact statements to investigate how emotional content changes across time periods and rating types.  A significant decrease in sentiment across time periods was observed, suggesting a slightly more negative tone in PolitiFact statements over time.   Moreover, statements associated with misinformation realized significantly lower sentiment than accurate and mixed-accuracy statements  (Figure~\ref{fig2}), which further supports research suggesting that misinformation relies on emotional content~\cite{35}.  However, there were also significant interactions that imply that statement sentiment is now a less reliable signal of misinformation than it was during the middle time period.  

Next, analysis was performed to discover common sources of misinformation and the consistency of those sources across time periods.  It was observed that online social networks (X, Facebook, Instagram, TikTok) and digital sources (Viral image, Bloggers) tend to be associated with high levels of misinformation and negative sentiment, whereas sources attributed to individual politicians tend to be more balanced across rating types and contain neutral or positive sentiment  (Figure~\ref{fig3}).    Moreover, analysis of source similarity across time showed that top PolitFact sources for both rating types have vastly changed since early time periods, but have become increasingly similar in recent time periods (Table ~\ref{table1}). 
 
Named-entity recognition was used to discover entities and entity labels that are common in misinformation, and their consistency across time periods. Analysis of entities contained in PolitiFact statements found that presidential incumbents and candidates are relatively more prevalent in statements containing misinformation, while US states tend to be present in accurate information (Figure~\ref{fig4}).  Analysis of entity similarity across time showed entities associated with misinformation in the early time period had less similarity to those in other time periods,  and that entities associated with accurate information have increased in similarity following the early time period (Table ~\ref{table2}). 

Finally, entity labels associated with people and organizations are more common in misinformation, while accurate statements are more likely to contain numeric entity labels, such as percentages and dates. Note that the differences at the entity label level (Figure~\ref{fig5}) tend to be smaller than at the raw entity level (Figure~\ref{fig4}).  Therefore, using entity labels will likely result in decreased ability to detect misinformation, but leveraging entities will require additional processing due to the increased number of categorical features.  Entity labels for PolitiFact statements show reasonably high similarity between all time periods. This suggests that top PolitFact statement entity labels (Table ~\ref{table3}) are generally more consistent than top entities (Table ~\ref{table2}) across time periods, but as discussed above, may result in decreased ability to detect misinformation. 

Taken together, these findings provide insight into robust linguistic features that are promising for detecting misinformation across time periods.  More specifically, consistent features of misinformation include online and digital sources, and statements with entities containing presidential incumbents and candidates, whereas relying on statement sentiment and features at the entity label level may not be as effective. However, generalization of these findings to the original content is unclear since this study analyzed PolitiFact statements that summarize source material.  Despite this limitation, it is hoped that these insights will prove useful for model developers.   

\section*{Supporting Information}
\paragraph*{S1 File.}
\label{S1_File}
{\bf Analysis Files.}  Contains: 1) Python functions to scrape, prepare, save and analyze data; 2) Jupyter notebook with analysis script to reproduce all figures and analysis in this paper; and 3) Saved data files that were used for analysis. Contact author to obtain.



\newpage

\begin{thebibliography}{10}

\bibitem{1}
Loomba, S., et al.
\newblock {{M}easuring the Impact of Exposure to COVID-19 Vaccine Misinformation on Vaccine Intent in the UK and US }
\newblock   Nat Hum Behav, 2021, 5, 337–348.

\bibitem{2}
Neely, S.R., et al.
\newblock {{V}accine Hesitancy and Exposure to Misinformation: a Survey Analysis}
\newblock    J GEN INTERN MED 2022. 37, 179–187.

\bibitem{3}
Ternullo, S.
\newblock {{I}m Not Sure What to Believe: Media Distrust and Opinion Formation during the COVID-19 Pandemic}
\newblock   American Political Science Review. 2022,116(3),1096-1109.

\bibitem{4}
Verma, G, Bhardwaj, A, Aledavood, T.
\newblock {{E}xamining the impact of sharing COVID-19 misinformation online on mental health}
\newblock   Sci Rep. 2022. 12, 8045

\bibitem{5}
Kavanagh, J, \&  Rich, MD.
\newblock {{T}ruth Decay: An initial exploration of the diminishing role of facts and analysis in American public life.}.
\newblock   RAND Corporation. 2018. ISBN: 978-0-8330-9994-5.

\bibitem{6}
Vosoughi, S., Roy, D., and Aral, S. 
\newblock {{T}he spread of true and false news online. }
\newblock    Science. 2018, 359, 1146-1151.

\bibitem{7}
Del Vicario, et al. 
\newblock {{T}he spreading of misinformation online. }
\newblock    PNAS. 2016, 113 (3) 554-559.

\bibitem{8}
Ecker, U.K.H., et al. 
\newblock {{T}he psychological drivers of misinformation belief and its resistance to correction.}
\newblock   Nature Reviews Psychology. 2022, 1, 13-29.

\bibitem{9}
Ceylan, G., Anderson, I. and Wood, W.  
\newblock {{S}haring of misinformation is habitual, not just lazy or biased.}
\newblock   PNAS. 2023, 120 (4).

\bibitem{10}
Ceylan, G., Anderson, I. and Wood, W.  
\newblock {{S}pread of misinformation on social media: What contributes to it and how to combat it.}
\newblock   Computers in Human Behavior. 2023, 141.

\bibitem{11}
Bateman, J. and Jackson, D.
\newblock {{C}ountering Disinformation Effectively: An Evidence-Based Policy Guide.}
\newblock   Carnegie Endowment for International Peace, 2024. 

\bibitem{12}
Chapp, C. and Aehl, P.
\newblock {{N}ewspapers and Political Participation: The Relationship Between Ballot Rolloff and Local Newspaper Circulation.}
\newblock   Newspaper Research Journal, 2021, 42, 2.

\bibitem{13}
Stearns, J. and Schmidt, C.
\newblock {{H}ow We Know Journalism Is Good for Democracy.}
\newblock   Democracy Fund Report, 2022.

\bibitem{14}
Dame, T. and Adjin-Tettey.
\newblock {{C}ombating Fake News, Disinformation, and Misinformation: Experimental Evidence for Media Literacy Education.}
\newblock   Cogent Arts \& Humanities, 2022, 9(1).

\bibitem{15}
Roozenbeek, J. and Sander van der Linden.
\newblock {{F}ake News Game Confers Psychological Resistance Against Online Misinformation.}
\newblock   Palgrave Communications, 2019, 5, 65.

\bibitem{16}
Roozenbeek, J. and Sander van der Linden.
\newblock {{A}lgorithmic Content Moderation: Technical and Political Challenges in the Automation of Platform Governance.}
\newblock   Big Data \& Society, 2020, 7(1).

\bibitem{17}
DiResta, R.
\newblock {{U}p Next: A Better Recommendation System.}
\newblock   Wired, 2018.

\bibitem{18}
Silverman, C.
\newblock {{A}s Facebook Abandons Fact-Checking, It’s Also Offering Bonuses for Viral Content.}
\newblock   Propublica, 2025.

\bibitem{19}
Roeder, A.
\newblock {{M}eta’s fact-checking changes raise concerns about spread of science misinformation}
\newblock   Harvard School of Public Health, 2025.

\bibitem{20}
Porter, E. and Wood, T.
\newblock {{T}he Global Effectiveness of Fact-Checking: Evidence From Simultaneous Experiments in Argentina, Nigeria, South Africa, and the United Kingdom}
\newblock   PNAS, 2021, 118 (37).

\bibitem{21}
Rich, P.R. and Zaragoza, M.S.
\newblock {{C}orrecting Misinformation in News Stories: An Investigation of Correction Timing and Correction Durability }
\newblock   Journal of Applied Research in Memory and Cognition, 2020, 3.

\bibitem{22}
Yadav, K.
\newblock {{P}latform Interventions: How Social Media Counters Influence Operations}
\newblock  Carnegie Endowment for International Peace, 2021.

\bibitem{23}
Stelter, B.
\newblock {{F}acebook to Start Putting Warning Labels on Fake News }
\newblock CNN, 2016.

\bibitem{24}
NewsGuard
\newblock {{R}ating Process and Criteria}
\newblock NewsGuard, 2023.

\bibitem{25}
Aslett, K., et al.
\newblock {{N}ews Credibility Labels Have Limited Average Effects on News Diet Quality and Fail to Reduce Misperceptions }
\newblock Science Advances, 2022, 8, 18.

\bibitem{26}
Lei, Z., et al.
\newblock {{B}IC: Twitter Bot Detection with Text-Graph Interaction and Semantic Consistency}
\newblock Proceedings of the 61st Annual Meeting of the Association for Computational Linguistics, 2023, 10326–10340.

\bibitem{27}
Knauth, J.
\newblock {{L}anguage-Agnostic Twitter-Bot Detection }
\newblock Proceedings of the International Conference on Recent Advances in Natural Language Processing, 2019, 550–558.

\bibitem{28}
Podorozhniak , A., et al.
\newblock {{R}esearch Application of the Spam Filtering and Spammer Detection Algorithms on Social Media }
\newblock Advanced Information Systems, 2023, 7(3), 60–66.

\bibitem{29}
Zheng , X., et al.
\newblock {{D}etecting spammers on social networks}
\newblock Neurocomputing, 2015, 159, 27-34.

\bibitem{30}
Roy, P.K. and Chahar, S.
\newblock {{F}ake Profile Detection on Social Networking Websites: A Comprehensive Review}
\newblock IEEE Transactions on Artificial Intelligence, 2020, 1(3), 271-285.

\bibitem{31}
Ramalingam, D. and Chahar, S.
\newblock {{F}ake profile detection techniques in large-scale online social networks: A comprehensive review}
\newblock Computers \& Electrical Engineering, 2018, 65, 165-177.

\bibitem{32}
Burdisso, S., et al.
\newblock {{R}eliability Estimation of News Media Sources: Birds of a Feather Flock Together}
\newblock NAACL, 2024.

\bibitem{33}
Ali Shaik, M., et al.
\newblock {{F}ake News Detection using NLP}
\newblock International Conference on Innovative Data Communication Technologies and Application, 2023, 399-405.

\bibitem{34}
Ali Shaik, M., et al.
\newblock {{A} Survey on Natural Language Processing for Fake News Detection}
\newblock Proceedings of the 12th Conference on Language Resources and Evaluation, 2020, 6086–6093.

\bibitem{35}
Carrasco-Farré, C. 
\newblock {{T}he fingerprints of misinformation: how deceptive content differs from reliable sources in terms of cognitive effort and appeal to emotions. }
\newblock Humanities and Social Sciences Communications, 2022, 9, 162.

\bibitem{36}
Carrasco-Farré, C. 
\newblock {{S}entiment Analysis for Fake News Detection. }
\newblock Electronics, 2021; 10(11).

\bibitem{37}
Bhutani, B., et al.
\newblock {{F}ake News Detection Using Sentiment Analysis. }
\newblock Twelfth International Conference on Contemporary Computing, 2019, 1-5.

\bibitem{38}
Bhutani, B., et al.
\newblock {{F}ake News Detection Using BERT Model with Joint Learning. }
\newblock  Arab J Sci Eng, 2021, 46, 9115–9127.

\bibitem{39}
Tsai, C.M.
\newblock {{S}tylometric Fake News Detection Based on Natural Language Processing Using Named Entity Recognition: In-Domain and Cross-Domain Analysis. }
\newblock  Electronic, 2023; 12(17).

\bibitem{40}
Allcott, H., Gentzkow, M. and Yu, C. 
\newblock {{T}rends in the diffusion of misinformation on social media. }
\newblock  Research and Politics, 2019, 10, 1-8.

\bibitem{40b}
Mimura, M., Ishimaru, T. 
\newblock {{A}nalyzing common lexical features of fake news using multi-head attention weights. }
\newblock  Interenet of Things, 2024, 28.

\bibitem{41}
Zhou, J, et al. 
\newblock {{S}ynthetic Lies: Understanding AI-Generated Misinformation and Evaluating Algorithmic and Human Solutions}
\newblock Proceedings of the 2023 CHI Conference on Human Factors in Computing Systems. 2023. 436, 1-20.

\bibitem{42}
Schlicht, E.J.
\newblock {{C}haracteristics of political misinformation over the past decade}
\newblock  BEA Research Symposium: The Impact of Disinformation and Misinformation on a Democratic Society. 2024.

\bibitem{43}
Islam, M.S.
\newblock {{C}OVID-19–Related Infodemic and Its Impact on Public Health: A Global Social Media Analysis.}
\newblock  Am J Trop Med Hyg. 2020, 103(4), 1621-1629

\bibitem{44}
Hutto, C., and Gilbert, E. 
\newblock {{V}ADER: A Parsimonious Rule-Based Model for Sentiment Analysis of Social Media Text. }
\newblock Proceedings of the International AAAI Conference on Web and Social Media. 2014, 8(1), 216-225.

\bibitem{45}
Wobbrock, J.O., et al.
\newblock {{T}he aligned rank transform for nonparametric factorial analyses using only ANOVA procedures. }
\newblock Association for Computing Machinery. 2011.

\bibitem{46}
Webber, W., et al.
\newblock {{A} similarity measure for indefinite rankings. }
\newblock Association for Computing Machinery. 2010, 28, 4, 1046-8188.



\end{thebibliography}
\end{document}